\documentclass[conference]{IEEEtran}
\IEEEoverridecommandlockouts

\usepackage{cite}
\usepackage{amsmath,amssymb,amsfonts}
\usepackage{algorithmic}
\usepackage{graphicx}
\usepackage{textcomp}
\usepackage{multirow}
\usepackage{booktabs}

\usepackage{xcolor}
\def\BibTeX{{\rm B\kern-.05em{\sc i\kern-.025em b}\kern-.08em
		T\kern-.1667em\lower.7ex\hbox{E}\kern-.125emX}}
\begin{document}
	
	\title{Effective Attention-Guided Multi-Scale Medical Network for Skin Lesion Segmentation\\

	}
	
	\author{\IEEEauthorblockN{Siyu Wang}
		\IEEEauthorblockA{\textit{Computer Science and Technology} \\
			\textit{Shandong Technology and Business University} \\
			Yantai, China \\
			wsy928057150@163.com}

		\and
		\IEEEauthorblockN{Hua Wang}
		\IEEEauthorblockA{\textit{Computer and Artificial Intelligence} \\
			\textit{Ludong University}\\
			Yantai, China  \\
			hwa229@163.com}
			
		\and
		\IEEEauthorblockN{Huiyu Li}
		\IEEEauthorblockA{\textit{Management Science and Engineering} \\
			\textit{Shandong University of Finance and Economics}\\
			Jinan, China \\
			huiyuroy@163.com}

		\and
		\IEEEauthorblockN{*Fan Zhang}
		\IEEEauthorblockA{\textit{Computer Science and Technology} \\
			\textit{Shandong Technology and Business University}\\
			Yantai, China \\
			zhangfan51@sina.com}

		}
	
	\maketitle
	
	\begin{abstract}
		In the field of healthcare, precise skin lesion segmentation is crucial for the early detection and accurate diagnosis of skin diseases. Despite significant advances in deep learning for image processing, existing methods have yet to effectively address the challenges of irregular lesion shapes and low contrast. To address these issues, this paper proposes an innovative encoder-decoder network architecture based on multi-scale residual structures, capable of extracting rich feature information from different receptive fields to effectively identify lesion areas. By introducing a Multi-Resolution Multi-Channel Fusion (MRCF) module, our method captures cross-scale features, enhancing the clarity and accuracy of the extracted information. Furthermore, we propose a Cross-Mix Attention Module (CMAM), which redefines the attention scope and dynamically calculates weights across multiple contexts, thus improving the flexibility and depth of feature capture and enabling deeper exploration of subtle features. To overcome the information loss caused by skip connections in traditional U-Net, an External Attention Bridge (EAB) is introduced, facilitating the effective utilization of information in the decoder and compensating for the loss during upsampling. Extensive experimental evaluations on several skin lesion segmentation datasets demonstrate that the proposed model significantly outperforms existing transformer and convolutional neural network-based models, showcasing exceptional segmentation accuracy and robustness.
	\end{abstract}
	
	\begin{IEEEkeywords}
		Medical Image Segmentation, Multi-Scale Feature Fusion, Attention Mechanism, Lightweight Network
	\end{IEEEkeywords}
	
	\section{Introduction}
	
Medical image segmentation plays a crucial role in extracting regions of interest, such as biological structures or pathological areas, which aids in diagnosis, treatment, and disease monitoring. With advancements in CT, MRI, and skin lesion imaging, the complexity of images has increased, making traditional methods that rely on manually designed features struggle to handle complex lesions, limiting accuracy and efficiency\cite{yao2024swift}, while recent generative and diffusion-based frameworks such as MoleBridge~\cite{zhangmolebridge} have demonstrated the potential of stochastic modeling for more controllable representation learning. In recent years, deep learning ~\cite{wang2025medical,xiao2025diffusion,yao2023ndc,zhang2024synergistic} has greatly advanced medical image segmentation. The U-Net architecture~\cite{ronneberger2015u} achieves efficient feature fusion through skip connections and has been widely adopted across diverse segmentation tasks. Its success has inspired numerous variants, such as Attention U-Net~\cite{oktay2018attention}, CA-Net~\cite{gu2020net}, and others. These methods enhance feature extraction capabilities through attention and context mechanisms. The success of Transformer in natural language processing~\cite{vaswani2017attention} has driven its application to visual tasks~\cite{zhang2023multi,zhang2024cf}. Vision Transformer~\cite{dosovitskiy2020image} has further advanced medical image segmentation, with models like Swin-Unet~\cite{cao2022swin} and TransUNet~\cite{chen2021transunet} using U-shaped Swin Transformer architectures for binary segmentation \cite{wang2025c3}. 

As shown in Figure \ref{fig:PHISIC}, despite progress, existing methods still face bottlenecks when handling complex lesions. Some models, despite using dilated convolutions or multi-scale paths \cite{jia2025rtgmff}, still fail to effectively capture fine-grained structures and global context. Additionally, traditional skip connections often use simple concatenation or addition, leading to information loss or blurred boundaries during upsampling, which impacts segmentation accuracy. To address these challenges, we propose EAM-Net, an attention-guided multi-scale medical segmentation network that integrates three key modules: the cross-mix attention module, the external attention bridge filters, and the multi-resolution multi-channel fusion block. To summarize, the main contributions of this paper can be described as follows:
	
	\begin{figure}[h]
		\centering
		\includegraphics[width=0.48\textwidth]{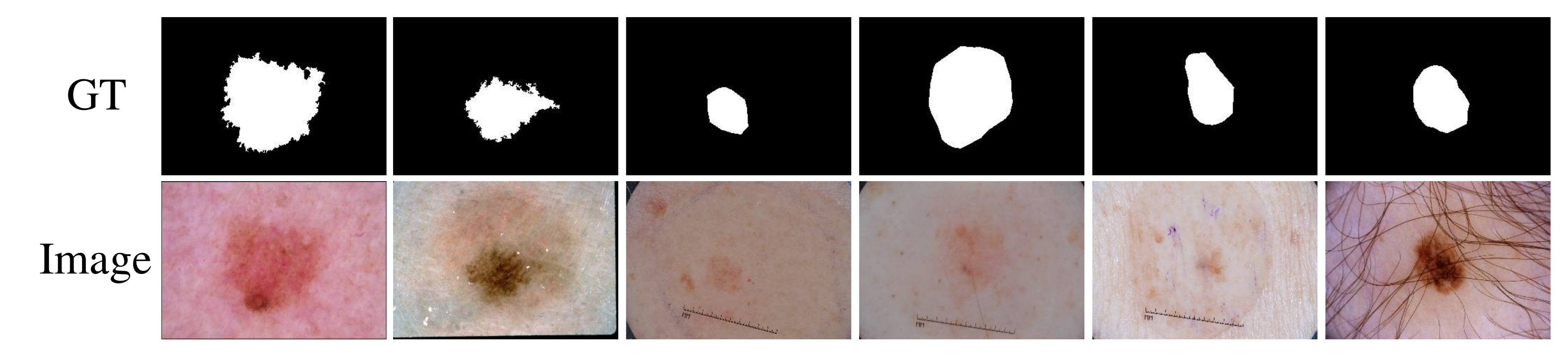}
		\caption{
			Some typical challenging skin lesions in the ISIC2018 and PH2 datasets. The first two are irregular lesions, the third is small lesions, the fourth and fifth are lesions with low contrast to the background, and the last is lesions with hair blocked.}
		\label{fig:PHISIC}
	\end{figure}

	\begin{itemize}
		\item The CMAM is proposed, which expands the scope of attention by integrating the $Q$, $K$, and $V$ representations of spatial attention and channel attention, achieving cross-guidance of global and local information, thereby enhancing sensitivity to lesion boundaries and subtle regions.
		\item The EAB module is designed, which introduces an external memory mechanism between the encoder and decoder to selectively enhance and filter features, compensating for the information loss in the skip connections.
		\item The MRCF is constructed, which uses a multi-branch structure combined with different convolution kernels and dilated convolutions to extract richer semantic information from multiple receptive fields, enhancing the model's scale robustness.
		\item Extensive experiments were conducted to demonstrate the superior performance of the newly proposed EAM-Net.
	\end{itemize}

	\begin{figure*}[h]
		\centering
		\includegraphics[width=0.8\textwidth]{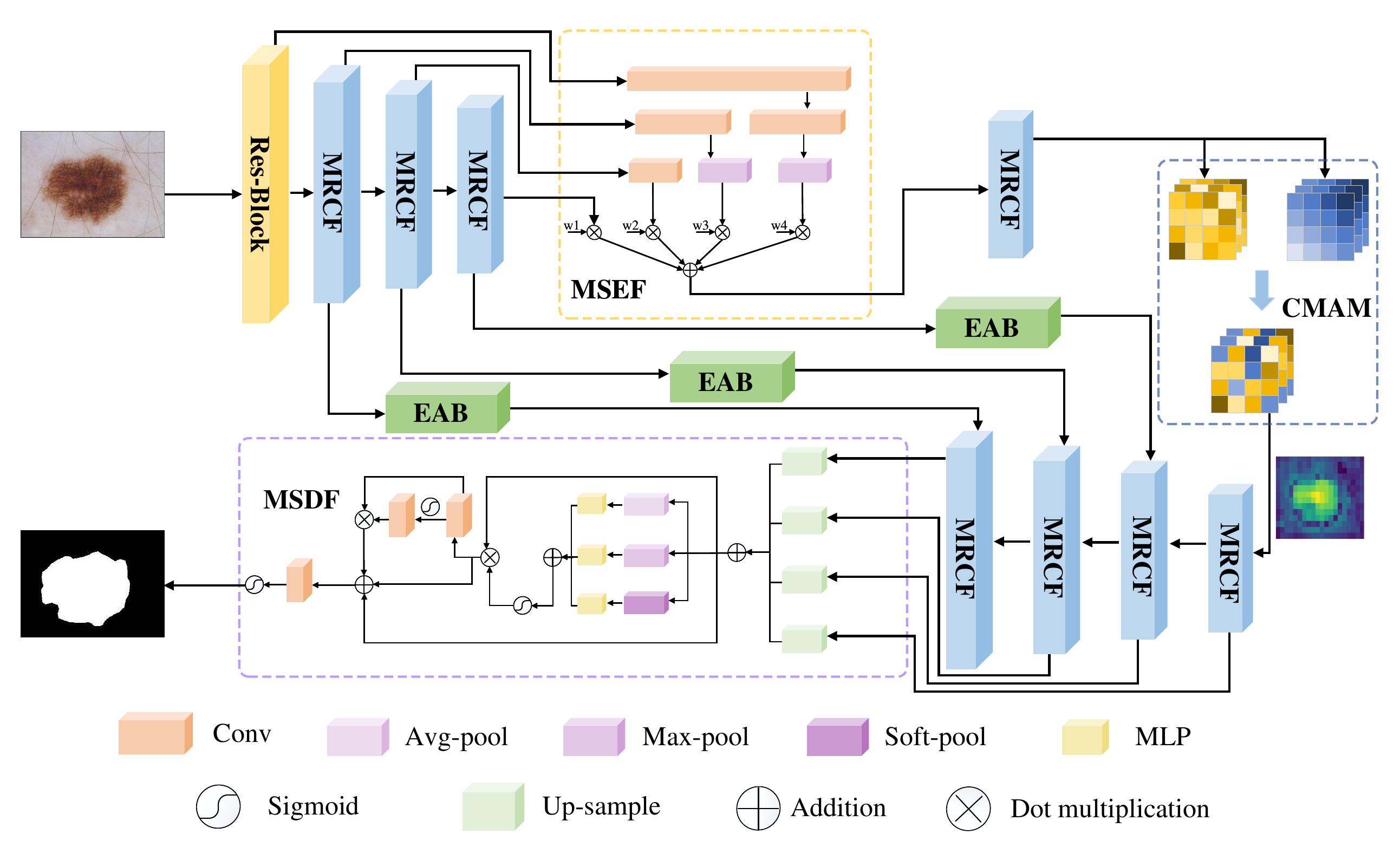}
		\caption{The main architecture of EAM-Net.The comments at the bottom represent the actual operations of the modules not labeled in the figure.}
		\label{fig:zhutu}
	\end{figure*}

\section{Proposed Method}
	
\subsection{Network Architecture}

This section presents the proposed network, with its architecture illustrated in Figure \ref{fig:zhutu}, and provides a detailed explanation of each module.

\subsection{Cross-Mix Attention Module}
	
In recent years, an increasing number of studies have adopted spatial and channel attention mechanisms for feature selection and localization~\cite{zhang2025enhancing}. However, most existing methods simply combine the two in a serial or parallel fashion, failing to fully exploit their complementary strengths. To overcome this limitation, we propose the CMAM, which performs cross-fusion between spatial and channel attention features to significantly expand the attention perception range. By computing context-aware weighting coefficients across multiple semantic spaces, CMAM enhances both the flexibility and precision of feature extraction. Moreover, it strengthens boundary modeling, enabling the network to more accurately capture complex and irregular lesion structures.

To enhance global semantic consistency and align local features with the CMAM. Given an input feature map $X \in \mathbb{R}^{H \times W \times C}$, the module applies spatial attention and channel attention separately to produce two enhanced feature maps $X^s$ and $X^c \in \mathbb{R}^{H \times W \times C}$. Here, $X^s$ is designated as the anchor feature, and $X^c$ as the complementary feature. Subsequently, the module computes the attention representations $Q$, $K$, and $V$ based on the following formulations to perform effective cross-contextual aggregation and feature fusion.

	\begin{equation}
		\begin{aligned}
			Q^s\ =\ X^s\ \cdot\ W_1^Q \\
			K^c\ =\ X^c\ \cdot\ W_1^K \\
			V^s\ =\ X^s\ \cdot\ W_1^V
		\end{aligned}
	\end{equation}
	
	where $W_1^Q$,$W_1^K$,$W_1^V\in{R}^{C\times D}$ represent the learned weight matrices, where $D$ is the number of channels after the linear transformation. The mixed attention was calculated using the following formula:
	\begin{equation}
		\begin{aligned}
			SA(Q^s,K^c,V^s)\ =\ SoftMax(Q^s\cdot{(K^c)}^T/\sqrt D){\cdot V}^s
		\end{aligned}
	\end{equation}
	
	where $SoftMax$ represents the row-wise SoftMax operation. Matching an attention query vector with another attention key vector, learning different feature mappings and similarity relationships respectively, is equivalent to introducing cross-information, improving the model's ability to understand complex structures, and guiding the global context to influence the attention weights of local features \cite{jia2025geodesic}.
	
	Similarly, $X^c$ as the anchor data and $X^s$ were chosen as the complementary data, and another set of $Q$, $K$, and $V$ was calculated:
	
\begin{equation}
	\begin{aligned}
		Q^c &= X^c \cdot W_2^Q \\
		K^s &= X^s \cdot W_2^K \\
		V^c &= X^c \cdot W_2^V
	\end{aligned}
\end{equation}

	The mixed attention for this instance can be calculated using the following formula:
	\begin{equation}
		\begin{aligned}
			CA(Q^c,K^s,V^c)\ =\ SoftMax(Q^c\cdot{(K^s)}^T/\sqrt D){\cdot V}^c
		\end{aligned}
	\end{equation}
	Therefore, the cross-mix attention module can be represented as follows:
	\begin{equation}
		\begin{aligned}
			CMAM\ =\ SA(Q^s,K^c,V^s)\ +\ CA(Q^c,K^s,V^c)
		\end{aligned}
	\end{equation}

	\subsection{Multi-Resolution Multi-Channel Fusion Module}
	
\begin{figure}[h]
		\centering
		\includegraphics[width=0.5\textwidth]{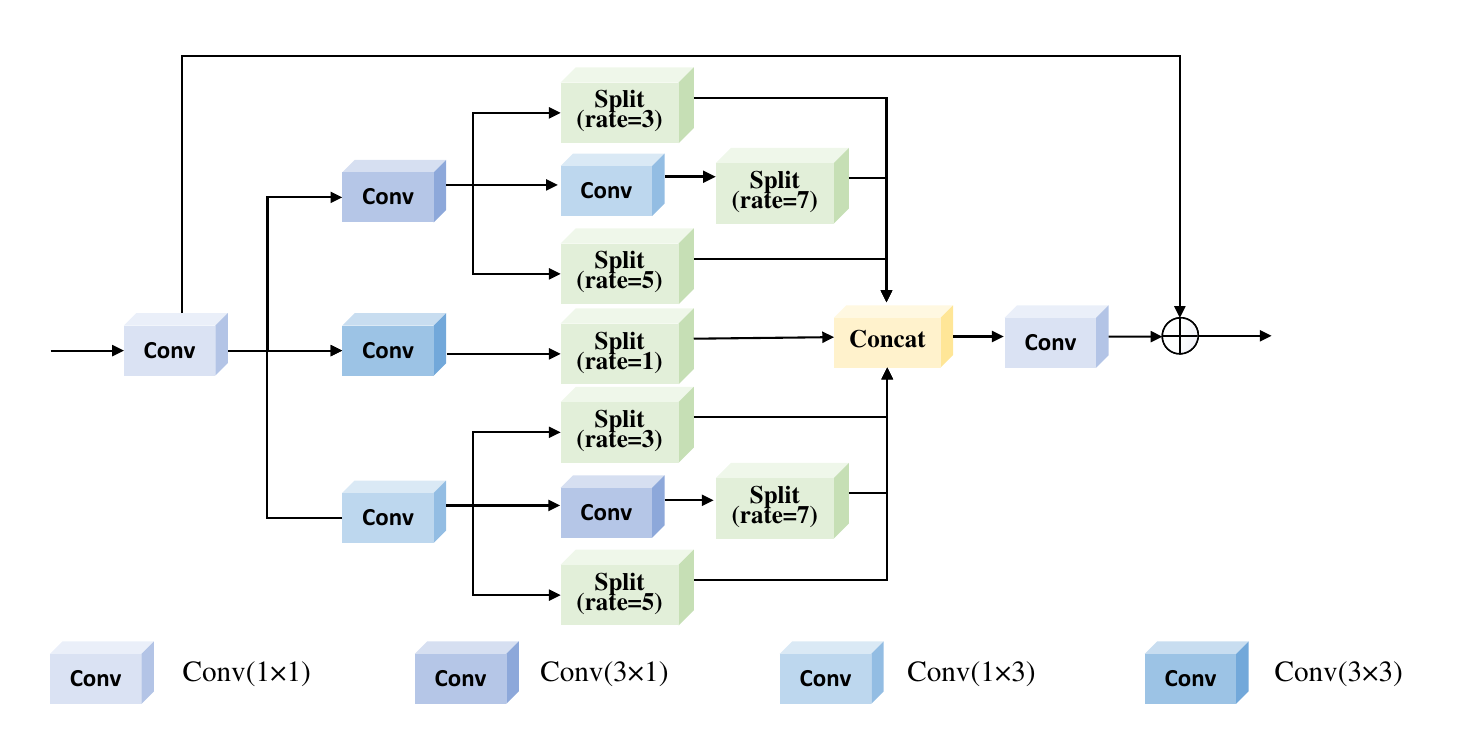}
		\caption{Main architecture of the MRCF module.}
		\label{fig:MRCFa}
	\end{figure}

\begin{figure}[h]
		\centering
		\includegraphics[width=0.5\textwidth]{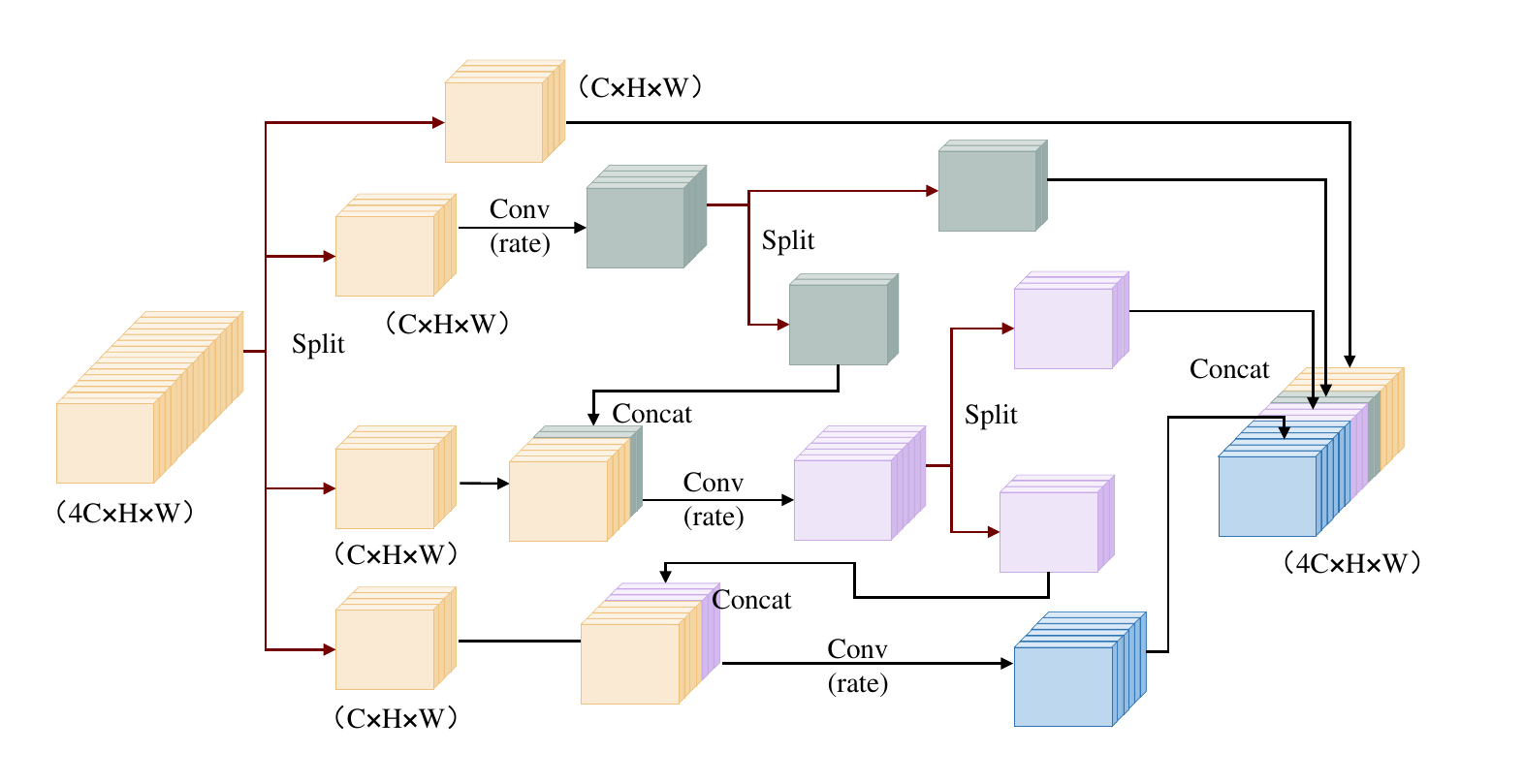}
		\caption{Implementation of the Split module in the MRCF module.}
		\label{fig:MRCFb}
	\end{figure}

To enhance the model's ability to represent complex lesion regions, we propose the MRCF module. The MRCF module consists of two core components: a multi-branch convolution unit with varying kernel sizes to simulate receptive fields at different scales, and a dilated convolution block to expand the receptive field without increasing computational cost. As shown in Figure~\ref{fig:MRCFa}, each branch starts with a 1×1 bottleneck convolution to reduce channel dimensionality. 3×3 convolutions replace 5×5 convolutions, and n×1 and 1×n convolutions are used to enhance non-linearity and model expressiveness. To improve cross-channel interaction, a dilated convolution structure is added to each branch,inspired by GhostNet \cite{han2020ghostnet}. The input features are split into four groups along the channel dimension: one group preserves raw information, and the remaining three groups undergo convolution at different scales to perform deep semantic encoding (see Figure~\ref{fig:MRCFb}), enhancing feature refinement.

Finally, the outputs from all branches are concatenated along the channel dimension and fused using a 1×1 convolution, producing a high-resolution, semantically rich feature map. The MRCF module offers a lightweight yet powerful multi-scale feature integration mechanism, improving the model's sensitivity to lesions of varying sizes and enhancing segmentation accuracy in complex medical images.

	\subsection{External Attention Bridge}
	
	\begin{figure}[h]
		\centering
		\includegraphics[width=0.5\textwidth]{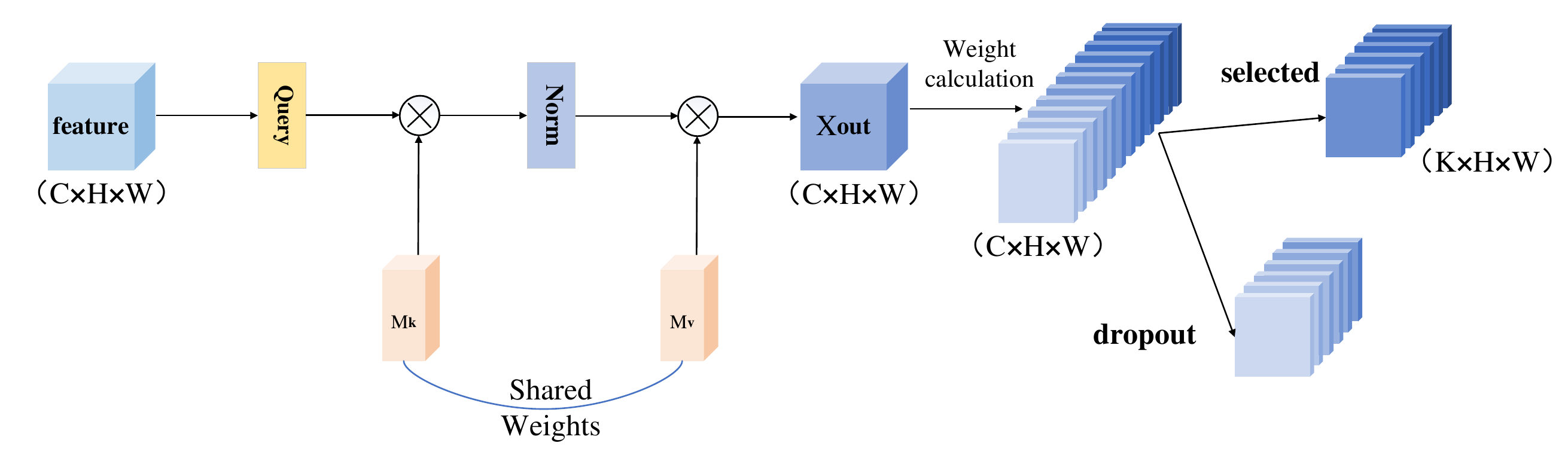}
		\caption{
			The architecture of the EAB module.}
		\label{fig:EAB}
	\end{figure}

The U-Net architecture incorporates skip connections linking the encoder and decoder. In most U-shaped networks \cite{dai2022ms}, skip connections typically use simple addition or concatenation, which limits their ability to fully utilize decoder information to address upsampling information loss. To solve this problem, an EAB was introduced. First, an external attention mechanism \cite{guo2022beyond} was used, as shown in the first half of Figure \ref{fig:EAB}. It was implemented with two cascaded linear and normalization layers. Essentially, it computes the attention between the input and external memory units $M_k$ and $M_v$. The memory units, denoted as ${M_k,M_v}\in{R}^{k\times n}$, can be expressed as follows:
	\begin{equation}
		\begin{aligned}
			&A=\left(\alpha\right)_{i,j}=Norm\left(FM_k^T\right) \\
			&F_{out}\ =\ AM_v
		\end{aligned}
	\end{equation}

	\begin{figure*}[t]
		\centering
		\includegraphics[width=0.8\textwidth]{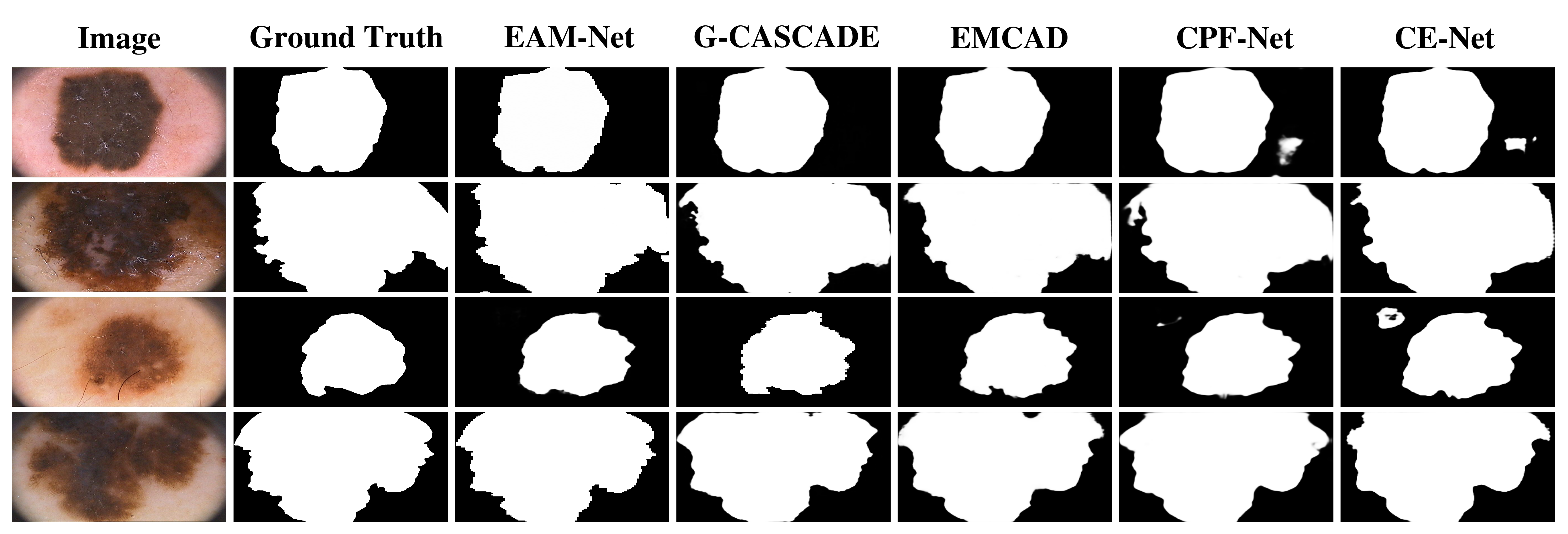}
		\caption{Comparison of visualization results between our EAM-Net and various competing models on the PH2 dataset.}
		\label{fig:PH2}
	\end{figure*}
	
	Where $n = H \ast W$, and $k$ is a hyperparameter. ${\left( \alpha  \right)_{i,j}}$ denotes the similarity between the $i$-th element and the $j$-th row of $M$. $F$ is the initial input, while $M_k$ and $M_v$ are input-independent memory units acting as global memory for the training dataset. This attention mechanism strengthens skip connections, enhancing the integration of global and local information, and mitigating the effects of information loss\cite{xiao2024confusion}. To select the necessary features, convolution operations and $L1$ normalization are used to select and concatenate the top $k$ channels with the maximum $L1$ norms for each batch. Specifically, the output of the external attention module, $x{out} \in \mathbb{R}^{H \times W \times C}$, is subjected to a convolution operation with weights $W$.
	\begin{equation}
		\begin{aligned}
			&out=\left|x_{out}\ast W\right|
		\end{aligned}
	\end{equation}
	\begin{equation}
		\begin{aligned}
			p_c\ =\ \sum_{i,j}{|{out}_{i,j,c}|}
		\end{aligned}
	\end{equation}
	\begin{equation}
		\begin{aligned}
			S=topk\left({p_{c,1},p_{c,2}\ldots,p_{c,C}}\ ,k\right)
		\end{aligned}
	\end{equation}
	\begin{equation}
		\begin{aligned}
			selected\ =\ {xS[1],xS[2],...,xS[k]}
		\end{aligned}
	\end{equation}
	
	where $p_c$ represents the $L1$ norm of channel $c$, $S$ represents the selected top-k channels, and $x_{S_{[c]}}$represents the portion corresponding to the c-th selected channel.

	\subsection{Multi-Scale Encoding Fusion module}
To further enhance the model’s ability to capture contextual dependencies during the encoding phase, the Multi-Scale Encoding Fusion (MSEF) module~\cite{dai2022ms} is incorporated into our framework. Unlike conventional multi-scale feature fusion approaches that simply concatenate or sum features across encoder stages, MSEF is designed to adaptively learn the importance weights of feature maps from different encoding levels. It operates in two steps: uniform rescaling and adaptive fusion, as shown in Figure \ref{fig:zhutu}.

	\subsection{Multi-Scale Decoding Fusion Module}
	A multi-scale decoding fusion module \cite{dai2022ms} is utilized to effectively integrate information from various encoder levels, ensuring more reliable segmentation results. This module combines the features from different levels using multiple pooling and attention mechanisms. Specifically, let $G_i(i\ \in\{1,2,3,4\})$ denote the feature representation of the i-th layer of the encoder, where $G_1$ has a resolution of 224 × 320, which matches the input resolution of the entire network. $G_i^1$ represents feature $G_i$ scaled to the size of the feature $G_1$, and $f_c^i$ denotes the compression of the scaled feature $G_i$ into four channels:
	\begin{equation}
		\begin{aligned}
			F_c\ =\ Concat(f_c^1(G_1),f_c^2(G_2),f_c^3(G_3),f_c^4(G_4))
		\end{aligned}
	\end{equation}
	By combining pooling with a multilayer perceptron, we obtain the coefficients $F_c$ for each channel. The scale coefficients attention vector $\alpha\in[0,1]^{4\times 1\times 1}$, where $F\cdot\alpha$ is used as a spatial attention block to generate spatial attention coefficients $\beta\in[0,1]^{1\times 224\times 320}$, where $\alpha\cdot\beta$ represents pixel-scale attention. The final output of the module is given by:
	\begin{equation}
		\begin{aligned}
			Y_{out}=\ F_c\cdot\alpha\cdot\beta+F_c\cdot\alpha+F_c
		\end{aligned}
	\end{equation}
	The specific structure is shown in Figure \ref{fig:zhutu}(MSDF).
	
	\begin{table*}[h]
		\centering
		\caption{Comparison of our EAM-Net with various competing models on the ISIC2018 and PH2 datasets, with the best results highlighted in bold.}
		\label{tab:1}
		\begin{tabular}{lcccccccccc}
			\hline
			\multirow{2}{*}{Networks} &\multirow{2}{*}{Para}& \multirow{2}{*}{Flops}& \multicolumn{4}{c}{ISIC2018} & \multicolumn{4}{c}{PH2} \\ \cmidrule(lr){4-7} \cmidrule(lr){8-11}
			& & & IoU   & Dice  & ACC   & Precision & IoU  & Dice  & ACC  & Precision \\ \hline
			U-Net        & 32.9M  & 65.53G  & 81.69 & 88.81 & 95.68 & 91.31     & 87.07 & 92.62 & 95.57 & 93.32   \\ 
			DenseASPP    & 33.7M  &107.74G  & 82.52 & 89.35 & 95.89 & 91.38     & 89.38 & 94.13 & 96.61 & 96.54   \\ 
			BCDU-Net     & 28.8M  &360.18G  & 80.84 & 88.33 & 95.48 & 89.68     & 87.41 & 93.06 & 95.61 & 95.56   \\ 
			CE-Net       & 29.0M  &  9.76G  & 82.82 & 89.59 & 95.97 & 90.67     & 89.62 & 94.36 & 96.68 & 94.54   \\ 
			CPF-Net      &30.65M  &  8.83G  & 82.92 & 89.63 & 96.02 & 90.71     & 89.91 & 94.52 & 96.72 & 95.56   \\ 
			PVT-CASCADE  &35.27M  &  8.20G  & 82.83 & 90.12 & 95.62 & 87.65     & 90.31 & 95.03 & 96.58 & 94.48   \\ 
			Parallel MERIT &147.9M & 33.43G  & 83.44 & 90.05 & 95.66 & 89.99     & 90.21 & 94.69 & 96.81 & 95.59   \\ 
			EMCAD        & 30.0M  &  6.36G  & 82.81 & 89.46 & 95.55 & 89.25     & 90.56 & 95.00 & 96.69 & 95.03   \\ 
			G-CASCADE    &141.4M  & 30.42G  & 82.98 & 89.57 & 95.74 & 89.73     & 90.11 & 94.75 & 96.21 & 94.94   \\ 
			\textbf{Ours} & 4.6M   & 16.85G  & \textbf{84.18} & \textbf{90.71} & \textbf{96.45} & \textbf{91.77} 
			& \textbf{90.88} & \textbf{95.15} & \textbf{96.99} & \textbf{96.58} \\ \hline
		\end{tabular}
	\end{table*}

	\section{Experiments}
	
	\subsection{Data Sets}
	To evaluate our model's performance, we used two publicly available skin lesion segmentation datasets: ISIC2018 \cite{codella2019skin} and PH2 \cite{mendoncca2013ph}. The ISIC2018 dataset contains 2,594 images of varying resolutions with ground truth labels across seven categories. The PH2 dataset, collected by the Dermatology Department of Pedro Hispano Hospital in Portugal, includes 200 8-bit RGB dermoscopic images with corresponding ground truth. For fair comparisons, both datasets were resampled to 224×320 pixels, as per prior studies \cite{dai2022ms}. ISIC2018 was divided into training, validation, and test sets with a 7:1:2 ratio, while PH2 was split into 80 training images, 20 validation images, and 100 testing images.

	\subsection{Experimental Settings}
	The proposed network was implemented in PyTorch, utilizing the Adam optimizer with an initial learning rate of 0.001 and a weight decay of 0.00005. A Cosine Annealing-WarmRestart learning rate schedule (T0 = 10, Tmult = 2) was applied. A combination of soft Dice loss \cite{bertels2019optimizing} and binary cross-entropy loss(BCE) are used as the network training loss functions. Training was conducted for 300 epochs on the ISIC2018 dataset and 350 epochs on the smaller PH2 dataset to ensure sufficient learning. All experiments were performed on a single NVIDIA A100 device.

	\subsection{Comparison with PH2 and ISIC2018}

Table \ref{tab:1} compares EAM-Net with other methods, evaluating performance in terms of parameters, IoU, Dice, ACC, and Precision. Classic methods like U-Net achieve high ACC but lower IoU, Dice, and Precision, with CPF-Net and BCDU-Net facing similar issues, often suffering from under- and over-segmentation in challenging cases, as shown in Figure \ref{fig:PH2}. While G-CASCADE and EMCAD achieve high Dice and ACC, their models are complex and struggle with boundary precision. In contrast, EAM-Net balances lightweight design with high IoU, Dice, ACC, and Precision, showing robustness and accuracy in difficult scenarios.

	\subsection{Ablation Study}
	
	To evaluate the effectiveness of the different modules in our proposed network, extensive ablation experiments were performed using the PH2 dataset. 
	
	\begin{table}[ht]
		\centering
		\caption{Ablation study of each module.}
		\label{tab:666}
		\resizebox{0.49\textwidth}{!}{
			\begin{tabular}{ccccccc}
				\hline
				\multicolumn{3}{c}{Networks}                         & \multirow{2}{*}{IoU} & \multirow{2}{*}{Dice} & \multirow{2}{*}{ACC}  & \multirow{2}{*}{Pre} \\ 
				\multicolumn{1}{c}{MRCF} & \multicolumn{1}{c}{CMAM} & \multicolumn{1}{c}{EAB}                          \\ \hline
				\multicolumn{1}{c}{}     & \multicolumn{1}{c}{}     &                          & 90.10                & 94.38                 & 96.47                                   & 94.90                       \\ 
				\multicolumn{1}{c}{$\checkmark$}     & \multicolumn{1}{l}{}     &                          & 90.22                 & 94.53                 & 96.49                                   & 95.45                      \\ 
				\multicolumn{1}{c}{}     & \multicolumn{1}{c}{$\checkmark$}     &                          & 90.32                 & 94.67                 & 96.53                                   & 96.12                      \\ 
				\multicolumn{1}{c}{}     & \multicolumn{1}{l}{}     & $\checkmark$                         & 90.28                 & 94.56                 & 96.48                                   & 95.74                      \\ 
				& \multicolumn{1}{c}{$\checkmark$}     &  \multicolumn{1}{c}{$\checkmark$}                        & 90.47                & 94.81                 & 96.60                                   & 96.22                      \\ 
				\multicolumn{1}{c}{$\checkmark$}     &      & \multicolumn{1}{c}{$\checkmark$}                          & 90.51                & 95.05                 & 96.87                                   & 95.96                      \\ 
				\multicolumn{1}{c}{$\checkmark$}     & \multicolumn{1}{c}{$\checkmark$}     &                          & 90.48                & 94.88                 & 96.67                                   & 96.29                      \\ 
				\multicolumn{1}{c}{$\checkmark$}     & \multicolumn{1}{c}{$\checkmark$}     &      $\checkmark$                    & 90.88                & 95.15                 & 96.99                                   & 96.58                      \\ \hline
			\end{tabular}
		}
	\end{table}

	\textbf{Ablation of Different Modules:} 
	Table \ref{tab:666} presents a quantitative analysis of the proposed methods, with single-module ablation studies showing that each method independently contributes to performance improvements. Notably, the standalone application of CMAM significantly enhanced segmentation performance, with increases of 0.22\%, 0.29\%, and 1.22\% in IoU, Dice, and Precision, respectively. Pairwise combination experiments further demonstrate that integrating multiple methods leads to superior performance, achieving more accurate segmentation results. These findings validate the complementarity of the proposed methods and highlight the exceptional performance of the network architecture in segmentation tasks.
	
	\begin{figure}[ht]
		\centering
		\includegraphics[width=0.5\textwidth]{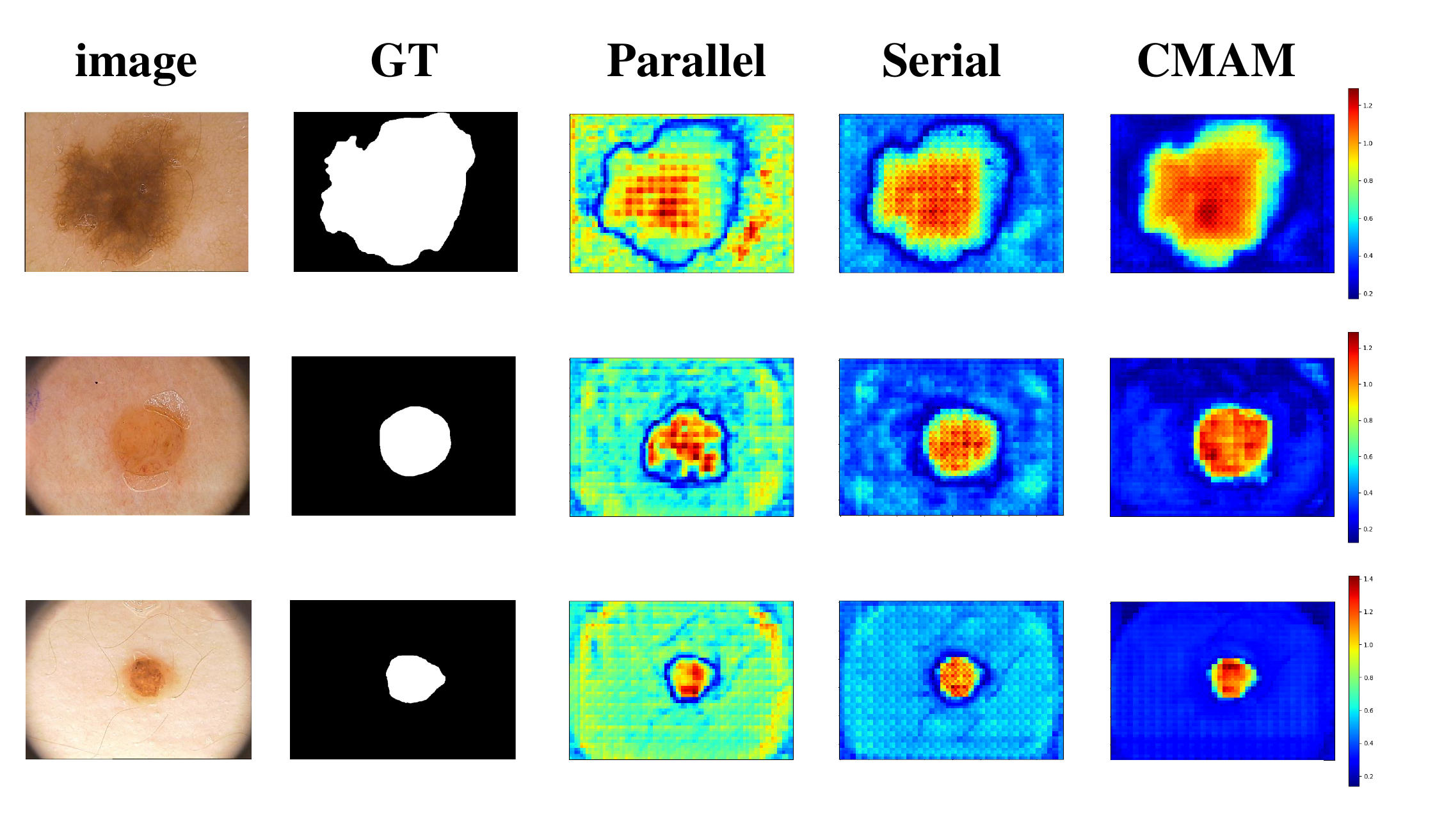}
		\caption{Comparison between our proposed CMAM and commonly used serial and parallel attention heatmaps.}
		\label{fig:compare}
	\end{figure}
	
	\textbf{Research on Attention Mechanism:} We propose in this paper a new attention CMAM. We visualize the weights learned by the attention module in the heat map to further compare and understand how our attention mechanism differs from other attention. It can be clearly seen from Figure \ref{fig:compare} that compared with simple serial and parallel attention, our proposed attention can better locate the skin lesion area, and capture clearer boundary information by combining local and global information.
	
\begin{figure}[ht]
		\centering
		\includegraphics[width=0.5\textwidth]{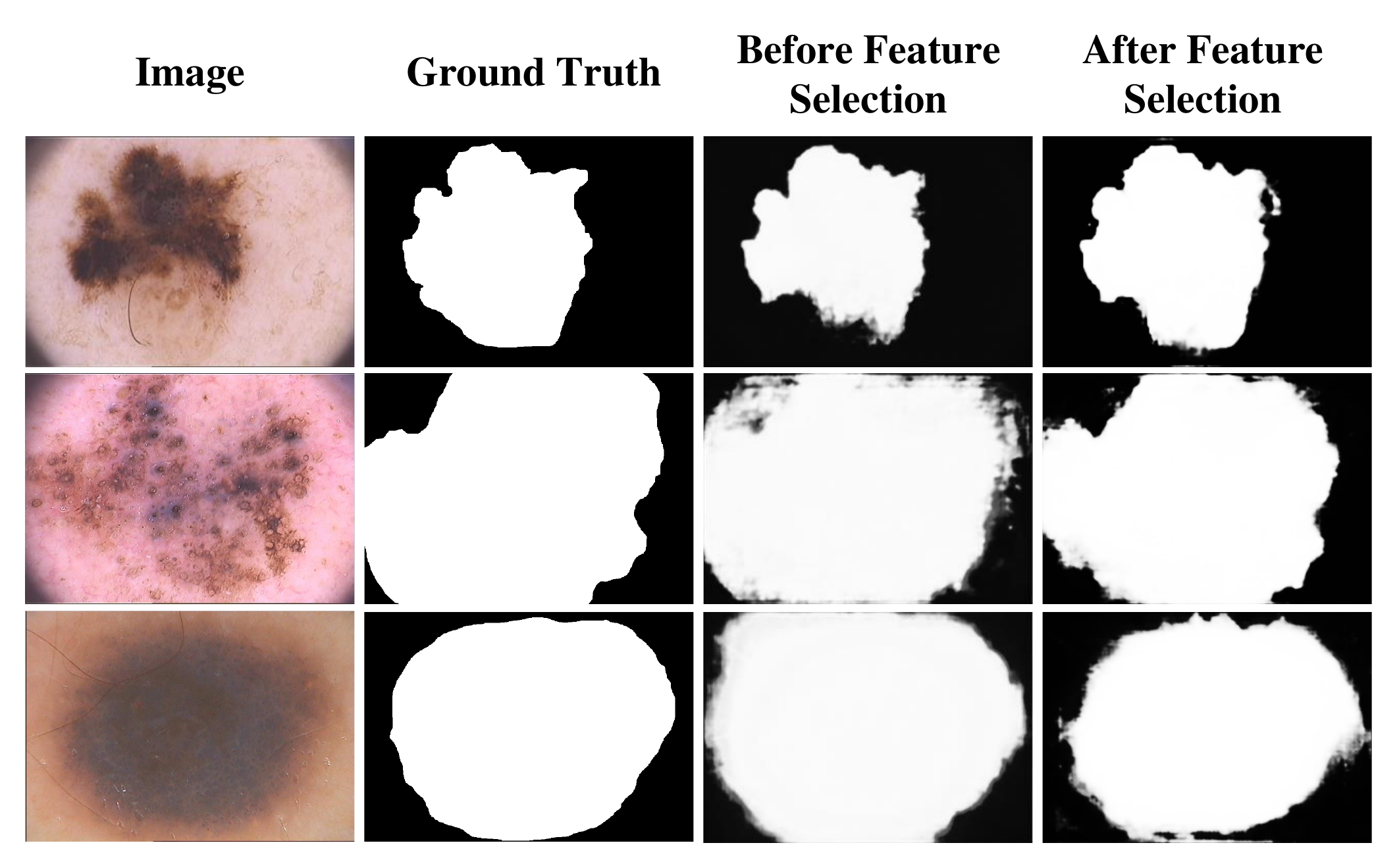}
		\caption{Comparison of the effect diagrams before and after feature selection clearly shows that the segmentation is clearer and accurate after feature selection.}
		\label{fig:EAB2}
	\end{figure}

	\textbf{Research within EAB:} We mentioned in the methodology above that feature filtering is added at the end of the EAB module. By observing Figure \ref{fig:EAB2}, the difference between before and after using feature screening can be clearly seen, which further confirms the importance of feature screening in improving the segmentation performance of the model. 
	
\section{Conclusion}
	
This study proposes an innovative multi-scale segmentation network aimed at addressing the challenges of irregular lesion shapes and low background contrast in skin lesion segmentation. By introducing the MRCF module, we effectively capture multi-receptive field information, enhancing sensitivity to lesions at different scales. The CMAM further optimizes the flexibility and accuracy of feature extraction. Additionally, the introduction of the EAB effectively mitigates the information loss caused by skip connections in the traditional U-Net model. These innovative designs work synergistically, enabling the network to achieve precise boundary detection and robust segmentation results when handling complex lesions. Experimental results show that, compared to existing mainstream methods, the proposed model demonstrates superior performance on multiple public skin lesion datasets. However, the model's focus on segmentation accuracy limits its consideration for real-world deployment and its application to other fields. Future work will explore generalizing the model to broader medical image analysis tasks and enhancing its practicality in real-world and clinical scenarios.

\section*{Acknowledgements}
This work was supported in part by the following: the National Natural Science Foundation of China under Grant Nos. U24A20219, 62272281, U24A20328, U22A2033, 62576193, the Special Funds for Taishan Scholars Project under Grant Nos. tsqn202306274, tsqn202507240, the Yantai Natural Science Foundation under Grant No. 2024JCYJ034, and the Youth Innovation Technology Project of Higher School in Shandong Province under Grant No. 2023KJ212.

	\bibliographystyle{IEEEtran}
	\bibliography{IEEEabrv,ref}
	
\end{document}